\documentclass[runningheads]{llncs}
\usepackage{graphicx}
\usepackage{array}
\usepackage{float}
\usepackage{array}
\usepackage{cite}
\usepackage{multirow}
\usepackage{booktabs}
\usepackage{colortbl} 
\usepackage{enumerate}
\usepackage{bbm}
\usepackage{amssymb}
\usepackage[backref]{hyperref} 

\usepackage{amsmath,graphicx}

\usepackage{amssymb}
\usepackage{hyperref}
\usepackage{cleveref}
\usepackage{marvosym}

\let\high\relax

\title{PANDA: Prompt-based Context- and Domain-aware Pretraining for Vision and Language Navigation}
\institute{}
\begin{document}

\author{Ting Liu \and Wansen Wu \and
Yue Hu\textsuperscript{(\Letter)} \and Youkai Wang \and Kai Xu \and Quanjun Yin
}

%
\institute{College of Systems Engineering, National University of Defense Technology, Changsha, China\\
\email{\{liuting20,huyue11,wuwansen14,wangyoukai,xukai09\}@nudt.edu.cn}}


%


%

\maketitle              
\begin{abstract}
Pretrained visual-language models have extensive world kno-
wledge and are widely used in visual and language navigation (VLN). However, they are not sensitive to indoor scenarios for VLN tasks. Another challenge for VLN is how the agent understands the contextual relations between actions on a path and performs cross-modal alignment sequentially. In this paper, we propose a novel Prompt-bAsed coNtext- and inDoor-Aware (PANDA) pretraining framework to address these problems. It performs prompting in two stages. In the indoor-aware stage, we apply an efficient tuning paradigm to learn deep visual prompts from an indoor dataset, in order to augment pretrained models with inductive biases towards indoor environments. This can enable more sample-efficient adaptation for VLN agents. Furthermore, in the context-aware stage, we design a set of hard context prompts to capture the \textit{sequence-level} semantics in the instruction. They enable further tuning of the pretrained models via contrastive learning. Experimental results on both R2R and REVERIE show the superiority of PANDA compared to existing state-of-the-art methods.

\end{abstract}

\section{Introduction}
Creating intelligent agents that can follow human instructions continues to be a major challenge in embodied artificial intelligence~\cite{das2018embodied}.
In particular, Vision and Language Navigation (VLN)~\cite{qi2020reverie,majumdar2020improving}, which requires an agent to follow natural language instructions and make sequential decisions in a photo-realistic simulated environment.

In recent years, there has been a growing tendency for VLN methods~\cite{majumdar2020improving,DBLP:conf/cvpr/HaoLLCG20,guhur2021airbert,DBLP:conf/cvpr/Lin0CLLL22} to build upon foundationally pretrained vision-and-language models. However, the distributional statistics of the large-scale web-scraped dataset conventionally employed for such pre-training often diverge substantially from the indoor domain of VLN environments. This domain gap leads to a disconnect in the ability to understand VLN scenarios.
In the meantime, the nature of sequential decision-making under partially observable environments, which essentially differs VLN from other reasoning tasks, requires a VLN agent to align between the sequences of action descriptions and viewpoints. With the above in mind, we argue that most existing works suffer from two significant shortcomings:

\begin{figure}[ht]
\vspace{-0.2cm}
    \centering
    \includegraphics[width = 0.6\linewidth]{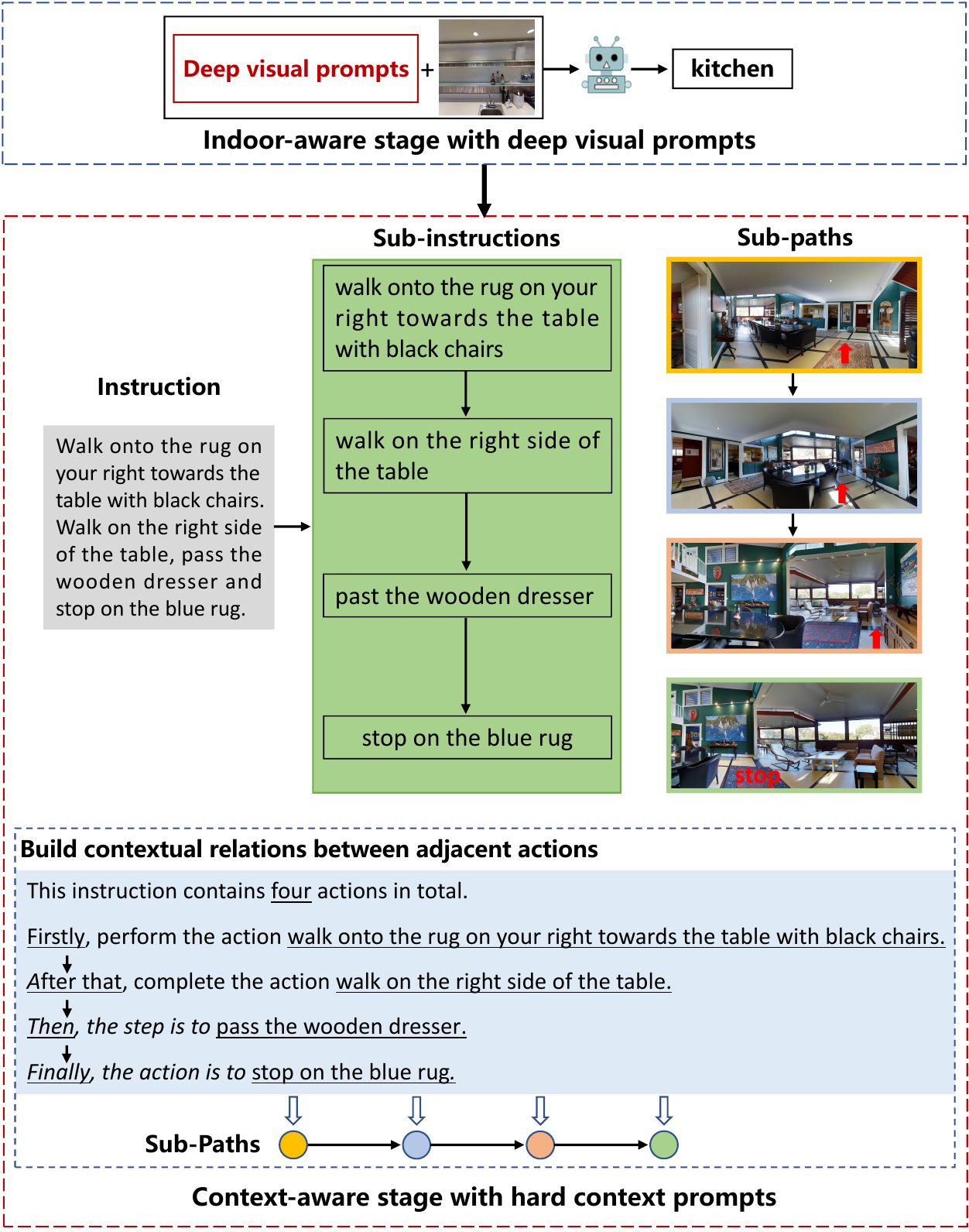}
     \caption{A simple demonstration of PANDA. In the indoor-aware stage, we exploit deep visual prompts learned from an indoor dataset by an auxiliary supervised task to help the VLN agent to abstract the indoor image semantics, then align sub-instructions with sub-paths sequentially by understanding the contextual relations between adjacent actions using hard context prompts in the context-aware stage.}
     \label{fig:fig1}

\end{figure}
\vspace{-0.6cm}

\begin{itemize}
    \item While pretrained models have powerful knowledge, their distributions diverge starkly from indoor navigation domains. Insensitive to indoor semantics, these models struggle to recapitulate the understanding requisite for dynamic reasoning within unfamiliar indoor scenes.
\end{itemize}
\vspace{-0.5cm}
\begin{itemize}
    \item Existing works ignore the contextual semantics implicitly embedded in the given instruction and the sequential relationships between fine-grained instructions (e.g., the instruction \textit{“Walk out of the bathroom and turn left”} can be divided into two ordinal sub-instructions, i.e., \textit{“first walk out of the bathroom”}, and \textit{“then turn left”}), and the agent may perform chaotic actions. 
\end{itemize}

Prompt engineering has been extensively studied and achieved significant success in NLP~\cite{DBLP:journals/corr/abs-2107-13586,DBLP:conf/emnlp/LesterAC21} and CV~\cite{radford2021learning,DBLP:journals/corr/abs-2109-11797}. Prompt engineering refers to the design of an input template that converts the expected output into a fill-in-the-blank format. With hand-designed prompts, the pioneer pretrained language model GPT-3\cite{DBLP:conf/nips/BrownMRSKDNSSAA20} has shown strong potential in the few-shot or zero-shot settings. CLIP~\cite{radford2021learning} embeds the text label of the object into the prompt template, which transforms the image recognition task into an image-text matching problem. CoOp~\cite{DBLP:journals/ijcv/ZhouYLL22} uses learnable vectors as text prompts to obtain the improvement of few-shot image classification. VPT~\cite{DBLP:conf/eccv/JiaTCCBHL22} introduces only a few learnable parameters into the input space while keeping the parameters of the pretrained model frozen during training. Prompt learning has also been introduced into VLN tasks~\cite{DBLP:conf/cvpr/Lin0CLLL22,liu2023dap} in some latest works. For example, our previous work DAP\cite{liu2023dap} applies a low-cost prompt tuning paradigm to learn shallow visual prompts for extracting in-domain image semantics. However, we found the shallow prompt lacks deep scene understanding.

Inspired by these works, we attempt to exploit prompt learning to solve the above problems and propose a novel Prompt-bAsed coNtext- and inDoor-Aware (PANDA) pretraining framework. A simple demonstration is shown in Figure \ref{fig:fig1}. PANDA improves pretrained general vision-language models in VLN with two stages: (i) In the indoor-aware stage, PANDA make a pretrained vision-language model aware of the specific VLN domain by adding deep visual prompts learned from indoor scene knowledge; (ii) In the context-aware stage, we aim to manually design context prompts to make the pretrained model aware of the contexts between navigational actions and reasoning about the entire sequence.

Specifically, to narrow the domain gap between the pretrained model and VLN tasks, we first generate a set of indoor datasets. Then we introduce a set of deep visual prompts in the input space of the visual encoder in a pretrained model. The aim is to enable the agent to identify the objects and scenes in the indoor images. Only deep visual prompts and an MLP head are learnable during training with the indoor datasets, while the parameters of the pretrained model are kept frozen. With prompt learning, deep visual prompts learned from the indoor dataset can adapt the pretrained models to VLN scenes very efficiently. In the context-aware stage, we first divide the R2R dataset~\cite{das2018embodied} into sub-instructions and sub-paths. Then, we use manually designed hard context prompts to explicitly align the predicted action sequence, which is characterized by a series of viewpoints, with the contextual information implicitly embedded in the given instruction, and instill both out-of-context and contextual knowledge in the instruction into cross-modal representations.  Contrastive learning is introduced to further tune the pretrained models for \textit{sequence-level} cross-modal alignment. 

In summary, the contributions of this work are summarized as: (i) We present PANDA to pretrain a representation model for VLN tasks that captures the indoor scene semantics and context semantics along the action sequence; (ii) We introduce deep visual prompts to adapt pretrained models to VLN tasks. Contrastive learning is also introduced to achieve effective alignment between textual prompts and visual semantics; (iii) PANDA shows promising performances and generalization ability with the help of prompt-based learning, and outperforms existing state-of-the-art methods.

\section{Related Works}
\noindent
\textbf{Vision and Language Navigation.}
Many methods~\cite{majumdar2020improving,DBLP:conf/cvpr/HaoLLCG20,guhur2021airbert} have been proposed for VLN, a famous embodied artificial intelligence task. Recently, the powerful representation abilities of pretrained models have attracted great attention. While the VLN-BERT\cite{majumdar2020improving} model was pretrained on a large set of web-crawled datasets to improve image-text matching, the PREVALENT \cite{hao2020towards} model is trained on a large amount of image-text-action triplets to learn generic representations of visual environments and language instructions. Following these works, a recurrent function~\cite{DBLP:conf/cvpr/Hong0QOG21} was introduced into the BERT model that significantly improves sequential action prediction. However, current VLN methods ignore the contextual information implicitly embedded in the given instruction and the sequential relations between sub-instructions. In addition, most pretrained models are trained on web-crawled general-purpose datasets, which incurs a considerable domain gap when used for VLN tasks.

\noindent
\textbf{Large-Scale Vision-Language Pretraining.}
Motivated by the success of the BERT model~\cite{DBLP:conf/naacl/DevlinCLT19} on NLP tasks, numerous vision-language pretrained (VLP) models~\cite{hao2020towards,li2020oscar} have been recently proposed. These VLP models have been applied to various vision-language tasks such as visual grounding~\cite{liu2023vgdiffzero} and vision and language navigation~\cite{hao2020towards}, etc., which have all made great performance breakthroughs. Despite their powerful visiolinguistic representation abilities, VLP models are not designed for tasks that entail sequential decision-making, such as VLN tasks. In this work, we aim to improve the VLP models to make them more suitable for VLN tasks.

\noindent
\textbf{Prompt Learning.}
The idea of prompt learning is to put the expected output as an unfilled blank into a prompt template that is incorporated into the input information, which has sparked significant interest in NLP~\cite{DBLP:journals/corr/abs-2107-13586,DBLP:conf/emnlp/LesterAC21}. There are two types of prompt templates: one is hard prompts designed manually, and the other is soft prompts learned automatically. For example, a cloze template \cite{DBLP:conf/emnlp/PetroniRRLBWM19} is designed manually to probe knowledge in pretrained language models can benefit many downstream tasks, and in P-tuning \cite{DBLP:journals/corr/abs-2103-10385}, deep prompt templates are learned in the continuous space by gradient descent without intricate design. Recent works~\cite{radford2021learning,zhou2022learning,DBLP:journals/corr/abs-2109-11797,tsimpoukelli2021multimodal} subsequently introduce prompt learning into the pretrained vision-language models. For example, CLIP~\cite{radford2021learning} embeds the text label of an image into a discrete template such as 
\textit{``A photo of a $\left\{object\right\}$"}, and the image recognition task can be transformed into an image-text matching problem.

\section{Method}

\begin{figure}[ht]
    \centering
    \includegraphics[width = 1\linewidth]{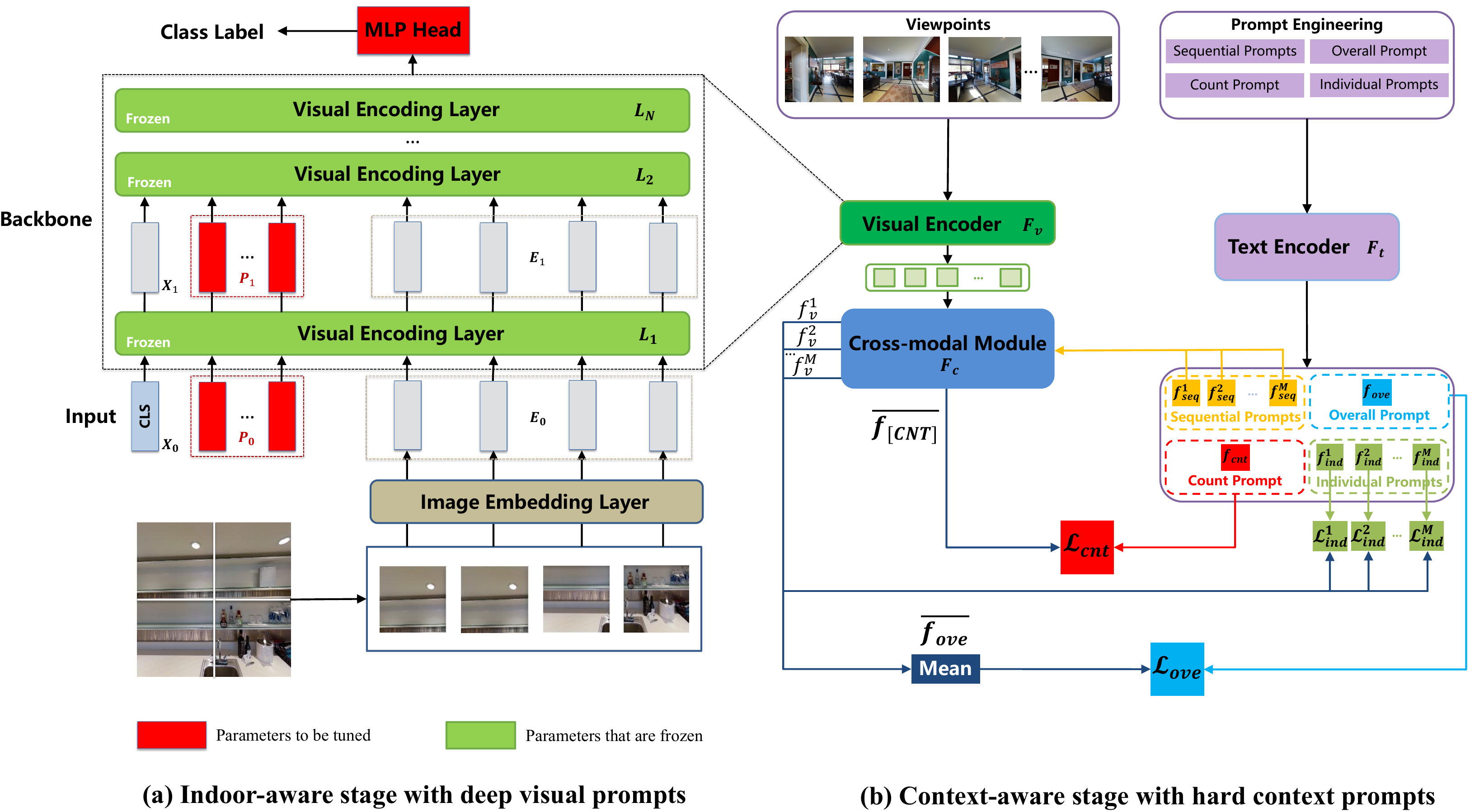}
     \caption{The overview of PANDA for VLN. We explore two forms of prompts: (a) deep visual prompts are inserted into the input space of the vision encoder, where only the parameters of deep visual prompts and the MLP head are updated during training; (b) we design hard context prompts to abstract the sub-instructions semantics and their sequential relations. The indoor-aware stage focuses on learning deep visual prompts to enhance the adaptation of backbone models to VLN tasks, while the context-aware stage aims to capture the contextual relations between actions on a trajectory and performs cross-modal alignment sequentially.}
     \label{fig:fig2}
\end{figure}

\subsection{VLN Problem Setup}

The VLN task can be expressed as follows: The VLN agent is put in a photorealistic environment such as Matterport3D \cite{DBLP:conf/3dim/ChangDFHNSSZZ17} simulator [8], it is assigned a random initial position and given a language instruction $I$ which contains $l$ word tokens. The VLN agent is required to find a route from the initial position to the target position. At each time step $t$, the agent observes the environment and makes a current action decision that updates the agent state $s_{t}$ to a new state $s_{t+1}$. The state includes historical information and the current spatial information consists of a viewpoint and orientation. All viewpoints are on the connectivity graph $G = \left\langle\\V, E\right\rangle$ of the environment \cite{DBLP:journals/corr/abs-1807-06757}, $V$ and $E$ represent navigable nodes and edges respectively. With the instructions, current state $s_{t}$ and visual observations $O_{t}$, the agent needs to execute the next actions $a_{t+1}$ one by one to navigate on the connectivity graph until stop at the target position.

\subsection{Prompt Engineering}

We introduce different forms of prompts in two stages for adapting pretrained vision-language models to VLN tasks in this subsection.

\noindent
\textbf{Learning Deep Visual Prompts Automatically.} In the indoor-aware stage, to narrow the domain gap,
we adopt the supervised learning method to learn deep visual prompts, taking indoor images as the input and the text of the corresponding object as labels. We introduce a set of continuous embeddings, i.e., deep visual prompts, in the input space after the input images are initially processed by the embedding layer. Deep visual prompts are automatically learned from an indoor dataset by prompt tuning, which helps the VLN agent to ground the object and scene descriptions in the instruction onto the visual perception. Deep visual prompts are learnable during training with the indoor dataset, while the parameters of pretrained models are kept frozen. Each visual prompt token is a learnable $d$-dimensional vector. As shown in Figure \ref{fig:fig2}(a), the form of deep visual prompts encoded by the $i\mbox{-}th$ visual encoding layer are continuous embeddings that can be represented as:
\begin{equation}
    P_{i} = \left\{p_{i}^{j}\in\mathbb{R}^d\mid j\in\mathbb{N},1\leq j \leq H\right\},
\end{equation}
 
\noindent
where $p_{i}^{j}$ represents the $j\mbox{-}th$ visual prompt in the $i\mbox{-}th$ layer and $H$ is the number of prompts. Note that besides $P_{i}$ ($ 1\leq i \leq N$), $P_{0}$ which is a part of the inputs to the visual encoder, takes the same form as $P_{i}$.

\noindent
\textbf{Designing Hard Context Prompts Manually.} In the context-aware stage, we aim to abstract the contextual semantics implicitly embedded in the given instruction and the semantics of sub-instructions in order. Recently, Bridge-Prompt \cite{li2022bridge}, the latest work in activity recognition from videos, discovers that human language is a powerful tool to depict the ordinal semantics between correlated actions. Motivated by this, we intend to manually design text prompts to realize the above purpose. Supposing that a sub-instruction (such as \textit{“walk into the hallway”}) constitutes a specific kind of agent action, we can use the prompt template as \textit{“perform the action {}”} (refers to the action description for the $i\mbox{-}th$ sub-instruction.) to abstract the semantics of the sub-instruction. However, if the sub-instruction is treated as a separate prompt instance, this out-of-context policy cannot describe the contextual semantics between adjacent ordinal sub-instructions. A more effective form of textual prompts should not only capture the semantics of individual sub-instructions, but also capture the contextual semantics and the sequential information between sub-instructions, and describe the overall semantics of the instruction. To this end, we manually design the hard context prompts consisting of four kinds of text prompts for VLN, as shown at the bottom of Figure \ref{fig:fig1}. We will empirically show the superiority of such a design over the exploitation of individual sub-instructions in Subsection \ref{Ablation Study}. Considering the instruction with $M$ sub-instructions:

\noindent
1) \textbf{A count prompt} abstracts the total number information of a sequence of actions contained in an instruction. We use the template as \textit{“This instruction contains $\left\{num\left(M\right)\right\}$ actions”} and denote the count prompt as $Y_{cnt}$.

\noindent
2) \textbf{A sequential prompt} abstracts the ordinal information for every sub-instruction. We use the template as \textit{“this is the $\left\{seq_{i}\right\}$ action”} and denote the sequential prompt as $y_{seq}^i$. The set of sequential prompts is as follows:
\begin{equation}
    Y_{seq} = \left[y_{seq}^1,\ldots,y_{seq}^M\right].
\end{equation}

\noindent
3) \textbf{An individual prompt} abstracts the semantic information of a sub-instruction. To integrate contextual information, we incorporate sequential information into the individual prompt and use the template as \textit{“$\left\{seq_{i}\right\}$, perform the action $\left\{a_{i}\right\}$”} for action $a_{i}$. We denote the individual prompt set as:
\begin{equation}
    Y_{ind} = \left[y_{ind}^1,\ldots,y_{ind}^M\right].
\end{equation}

\noindent
4) \textbf{An overall prompt} abstracts the overall information for the complete instruction. The overall prompt is made up of all individual prompts, which can be expressed as:
\begin{equation}
    Y_{ove} = Concat\left(y_{ind}^1,\ldots,y_{ind}^M\right).
\end{equation}

\subsection{Indoor-aware Stage with Deep Visual Prompts}

Deep visual prompts can be widely used in vision-language pretrained models to better understand indoor image semantics. We apply the pretrained model PREVALENT \cite{hao2020towards} for demonstration. We inject indoor visual knowledge into the visual encoder $F_{v}$ of the pretrained PREVALENT model by prompt tuning. 

\noindent
\textbf{Text Generation with the CLIP Model.} 
we take the powerful cross-modal pretrained model CLIP to automatically generates text labels corresponding to the indoor image from the Matterport3D \cite{DBLP:conf/3dim/ChangDFHNSSZZ17} dataset. The method takes full advantage of the knowledge learned from the CLIP~\cite{radford2021learning} and builds an indoor image-text dataset. We first encode the prompt template \textit{``A photo of a $\left\{object\right\}$"} by the text encoder of the CLIP, where the label represents object classes or rooms. Then we encode the image by the image encoder and calculate the similarity of the text embedding and image embedding. Finally, we choose the text with the highest matching score for the image. Through the above methods, we can automatically generate the indoor image-text dataset.

\noindent
\textbf{Deep Visual Prompt Tuning.} Deep visual prompts are inserted into the input space of the $N$-layer vision encoder in the PREVALENT. The output of the $i\mbox{-}th$ visual encoding layer is formulated as:

\begin{small}
    \begin{equation}
         \left[{\mathbf{X}}_{i},\mathbf{P}_{i},\mathbf{E}_{i}\right] = L_{i}\left(\mathbf{X}_{i-1},\mathbf{P}_{i-1},\mathbf{E}_{i-1}\right), \quad i=1,2,\ldots,N
    \end{equation}
\end{small}
\vspace{-0.2cm}

\noindent
where $\mathbf{X}_{i}$, $\mathbf{P}_{i}$, and $\mathbf{E}_{i}$ denote the $\left[CLS\right]$, prompts and image features respectively encoded by the $i\mbox{-}th$ visual encoding layer. The output $\mathbf{X}_{N}$ at the $N$-th layer of the visual encoder is mapped by an MLP head to a predicted class probability distribution $y$. 

The PREVALENT model is retrained on indoor image-text pairs that we have prepared, as shown in Figure \ref{fig:fig2}(a). Firstly, we freeze all parameters of the PREVALENT backbone model, which could not be updated during the training process. Then we add additional visual prompts on the first $n$ layers $\left(n\leq N\right)$, and an MLP head after the $N\mbox{-}th$ layer and deep visual prompts are learnable during training. We apply a cross-entropy loss to only optimize deep visual prompts and the linear head via gradients during prompt tuning. With such a low-consumption auxiliary classification task, the visual prompts are expected to inject the knowledge of object-level and scene-level indoor image semantics into the PREVALENT model.

\subsection{Context-aware Stage with Hard Context Prompts}

\noindent
\textbf{Sub-instructions and Sub-paths Generation.} In order to learn the ordinal relations between sub-instructions, and match a sub-instruction with its corresponding sub-path, we generate a fine-grained training dataset. We apply the FGR2R~\cite{DBLP:conf/emnlp/HongOWG20} method to divide the instructions into sub-instructions and pair each sub-instruction with its corresponding sub-path. Instructions are divided by \textit{“and”}, \textit{“comma”}, and \textit{“period”} delimiters. An illustrative example is provided here. We divide the given instruction \textit{“Walk onto the rug on your right towards the table with black chairs. Walk on the right side of the table, past the wooden dresser and stop on the blue rug.”} into \textit{“Walk onto the rug on your right towards the table with black chairs”}, \textit{“Walk on the right side of the table”}, \textit{“past the wooden dresser” and “stop on the blue rug”}, as shown in Figure \ref{fig:fig1}.

\noindent
\textbf{Fine-grained Alignment by Contrastive Learning.} The viewpoints along a path are first passed through the visual encoder $F_{v}$ updated in the indoor-aware pretraining stage to generate visual embeddings. We manually design the hard context prompts $\left(Y_{cnt},Y_{seq},Y_{ind},Y_{ove}\right)$ for the path. As shown in Figure \ref{fig:fig2}(b), the text encoder $F_{t}$ abstracts the embeddings of the hard context prompts as $\left(f_{cnt},f_{seq},f_{ind},f_{ove}\right)$, 
respectively. The visual embeddings and the sequential prompts embeddings are then passed through the cross-modal encoder $F_{c}$ to abstract the image features $f_{v}^{i}$ of the $i\mbox{-}th$ sub-path.

We input the $i\mbox{-}th$ sequential prompt feature $f_{seq}^i$ to the cross-modal module, which allows the cross-modal module to focus on the sequential information of each ordinal action. In addition, we add a learnable count token $\overline{f_{[CNT]}}$ in $F_{c}$ to extract quantitative information to match the count prompt ${f_{cnt}}$. 

Contrastive vision-text learning maximizes the similarity between encoded visual features and text features. We encode the sub-instruction $x$ and its corresponding sub-path $y$ with the text encoder and the visual encoder, respectively, generating text representation $r_{x}$ and vision representation $r_{y}$. The cosine similarity between $r_{x}$ and $r_{y}$ can be calculated as follows:
\begin{equation}
    s\left(r_{x},r_{y}\right) = \frac{r_{x}\cdot r_{y}}{\mid r_{x} \mid \mid r_{y} \mid }.
\end{equation}
For a batch of the text representation $R_{x}$ and the vision representation $R_{y}$, the batch similarity matrix $S$ can be denoted as:
\begin{equation}
    S\left(R_{x},R_{y}\right) = \left[\begin{array}{ccc}s\left(r_{x_{1}},r_{y_{1}}\right) & \cdots & s\left(r_{x_{1}},r_{y_{M}}\right)\\ \vdots & \ddots & \vdots \\s\left(r_{x_{M}},r_{y_{1}}\right) & \cdots & s\left(r_{x_{M}},r_{y_{M}}\right)\end{array} \right ],
\end{equation}

\noindent
where $M$ is the number of sub-instructions (sub-paths). We respectively apply a normalized function to the rows and columns on $S\left(R_{x}, R_{y}\right)$ to get $S_{V}\left(R_{x}, R_{y}\right)$ and $S_{T}\left(R_{x}, R_{y}\right)$. We assign the similarity score of positive pairs to 1 while negative pairs to 0, thus obtaining the batch similarity matrix $GT$ of ground truth. Our training objective is to maximize the similarity between the matrix $S$ and $GT$. We use the Kullback–Leibler divergence as the contrastive loss:
\begin{equation}
    D_{KL}(P\Vert Q) = \frac{1}{N^2} \sum_{i=1}^N \sum_{j=1}^N P_{ij} \log \frac{P_{ij}}{Q_{ij}},
\end{equation}

\noindent
where $P$ and $Q$ are $N\times N$ matrices. The contrastive loss for vision-text pairs can be defined as:
\begin{equation}
    \mathcal{L}=\frac{1}{2}\left[D_{KL}(S_T\Vert GT)+D_{KL}(S_V\Vert GT)\right].
\end{equation}

\noindent
The above formula is used to calculate the three parts of vision-text contrastive losses:

\begin{enumerate}[1)]
    \item $\mathcal{L}_{ind}^i$ is the contrastive loss between $f_{ind}^i$ and $f_{v}^i$, which allows the model to align each sub-instruction and the corresponding sub-path.
    \item $\mathcal{L}_{ove}$ is the contrastive loss between the mean-pooled $\overline{f_{ove}}$ and overall prompt feature $f_{ove}$, where $\overline{f_{ove}}$ is the mean-pooled features of all image features. $\mathcal{L}_{ove}$ captures the relationship between the overall instruction and the whole path.
    \item $\mathcal{L}_{cnt}$ is the contrastive loss between the learnable count token $\overline{f_{[CNT]}}$ and the count prompt feature ${f_{cnt}}$, which captures quantitative information of all actions.
\end{enumerate}

\noindent
The overall loss of the context prompt framework can be denoted as follows:
\begin{equation}
    \mathcal{L}=\lambda_1\mathcal{L}_{ove}+\lambda_2\mathcal{L}_{cnt}+\sum^{M}_{i=1}{\mathcal{L}^i_{ind}},
\end{equation}
where $\lambda_1$ and $\lambda_2$ are balance coefficients.

\section{Experiment}
\subsection{Experimental Setup}

\noindent
\textbf{Implementation Details.} Our training process includes updating the pretrained model and adapting it to downstream VLN tasks. Without loss of generality, our baseline agent follows the architecture of RecBERT~\cite{DBLP:conf/cvpr/Hong0QOG21}, which initializes from the pretrained model OSCAR~\cite{li2020oscar} learned from out-of domain datasets or PREVALENT \cite{hao2020towards} learned from VLN datasets. In the indoor-aware stage, we pretrain the PREVALENT model on our generated about 1000 indoor image-text pairs with prompt tuning for 20 epochs with batch size 10, and the number of deep visual prompts is 10. In the context-aware stage, we continue to train the PREVALENT model updated in the indoor-aware with pairs of sub-instruction and sub-paths for 20 epochs with batch size 20. After that, we adapt PANDA to the downstream generative VLN task with fine-tuning. Based on a simple parameter sweep, values of $\lambda_1$ and $\lambda_2$ are set to be 0.5 and 0.1, respectively. For R2R, we train the agent on the raw training data and the augmented data from PREVALENT for 300,000 iterations, and the batch size is 8. For REVERIE, we train the agent for 200,000 iterations with batch size 8. All experiments are conducted on a single NVIDIA 3090 GPU.

\definecolor{Gray}{gray}{0.94}
\begin{table*}[t!]
\begin{center}
\caption{Comparison with the SOTA methods on R2R dataset.}
\resizebox{0.70\textwidth}{!}{
\begin{tabular}{lrrrrrr>{\columncolor{Gray}}r>{\columncolor{Gray}}rrr>{\columncolor{Gray}}r>{\columncolor{Gray}}r}
    \toprule 
    \multicolumn{1}{c}{\multirow{2}{*}{Agent}} & \multicolumn{4}{c}{Val Seen} & \multicolumn{4}{c}{Val Unseen} & \multicolumn{4}{c}{Test Unseen} \\
    \cmidrule(r){2-5} \cmidrule(r){6-9} \cmidrule(r){10-13}
    \multicolumn{1}{c}{} & \multicolumn{1}{c}{TL} & \multicolumn{1}{c}{NE$\downarrow$} & \multicolumn{1}{c}{SR$\uparrow$} & \multicolumn{1}{c}{SPL$\uparrow$} & \multicolumn{1}{c}{TL} & \multicolumn{1}{c}{NE$\downarrow$} & \multicolumn{1}{c}{SR$\uparrow$} & \multicolumn{1}{c}{SPL$\uparrow$} & \multicolumn{1}{c}{TL} & \multicolumn{1}{c}{NE$\downarrow$} & \multicolumn{1}{c}{SR$\uparrow$} & \multicolumn{1}{c}{SPL$\uparrow$} \\
    \midrule
    Random                & 9.58  & 9.45 & 16 & -    & 9.77  & 9.23 & 16 & -    & 9.89  & 9.79 & 13 & 12 \\
    Human                 & -     & -    & -    & -    & -     & -    & -    & -    & 11.85 & 1.61 & 86 & 76 \\
    \midrule

    PRESS \cite{DBLP:conf/emnlp/LiLXBCGSC19}~
    & 10.57 & 4.39 & 58 & 55 & 10.36 & 5.28 & 49 & 45 & 10.77 & 5.49 & 49 & 45 \\
    EnvDrop \cite{DBLP:conf/naacl/TanYB19}~
    & 11.00 & 3.99 & 62 & 59 & 10.70 & 5.22 & 52 & 48 & 11.66 & 5.23 & 51 & 47 \\
    PREVALENT \cite{hao2020towards}~
    & 10.32 & 3.67 & 69 & 65 & 10.19 & 4.71 & 58 & 53 & 10.51 & 5.30 & 54 & 51 \\
    EnvDrop+REM \cite{DBLP:conf/iccv/0002ZCLGS21}~
    & 11.13 & 3.14 & 70 & 66 & 14.84 & 4.99 & 53 & 48 & 10.73 & 5.40 & 54 & 50 \\
     AuxRN \cite{zhu2020vision}~
    & -     & 3.33 & {70} & \ 67 & -     & 5.28 & 55 & 50 & -     & 5.15 & 55 & 51 \\
    ORIST \cite{DBLP:conf/iccv/QiPH0H021}~
    & - & - & - & - & 10.90 & 4.72 & 57 & 51 & 11.31 & 5.10 & 57 & 52 \\
    NvEM \cite{DBLP:conf/mm/AnQHWWT21}~  & 11.09 & 3.44 & 69 & 65 & 11.83 & {4.27} & {60} & {55} & 12.98 & {4.37} & {58} & {54} \\
    EnvDrop+SEvol \cite{DBLP:conf/cvpr/ChenGMZ022} & 12.55     &  3.70   &  61   &  57    &  14.67   &  4.39   &  59 &  53   &  14.30    &  \high{\textbf{3.70}}  &   59  &  55   \\  
    NvEM+SEvol \cite{DBLP:conf/cvpr/ChenGMZ022}
    & 11.97     &  3.56   &  67   &  63    &  12.26   &  \ 3.99   &  \ 62 &  57   &  13.40    &  \ 4.13  &   \ 62  &  \ 57   \\

    ProbES \cite{DBLP:conf/acl/LiangZLXL22}
    & 10.75     &  2.95   &  73   &  69    &  11.58   &  \ 4.03   &  \ 61 &  \ 55   &  12.43    &  \ 4.20  &   \ 62  &  \ 56   \\
     ADAPT \cite{DBLP:conf/cvpr/Lin0CLLL22}
    & 11.39     & 2.70   &  74   &  69    &  12.33   &   3.66   &   66 &   59   &  13.16    &  4.11 &   \ 63  &  \ 57   \\
     GRVLN-BERT \cite{DBLP:conf/swarm/ZhangQZZYWWX23}
    & 11.08     & 2.58   &  75   &  71    &  12.49   &   3.81   &   62 &   56   &  12.78    &  3.96 &   \ 63  &  \ 57   \\
    \hline
    RecBERT (init. OSCAR)~\cite{DBLP:conf/cvpr/Hong0QOG21}	& 10.79 & 3.11 & 71 & 67 & 11.86 & 4.29 & 59 & 53 & 12.34 & 4.59 & 57 & 53 \\
    RecBERT (init. PREVALENT) \cite{DBLP:conf/cvpr/Hong0QOG21} ~ &  11.13 & 2.90 & 72 & 68 & 12.01 & 3.93 & 63 & 57 & 12.35 & 4.09 & 63 & 57 \\
   
    \midrule
    PANDA(Ours)~ & \ 10.65 & \high{\textbf{2.54}} & \high{\textbf{75}} & \high{\textbf{72}} &  12.08 & \high{\textbf{3.50}} & \high{\textbf{66}} & \high{\textbf{60}} & 12.31 & {3.86} & \high{\textbf{64}} & \high{\textbf{59}} \\

    \bottomrule
\end{tabular}
}
\vspace{-0.5cm}
\label{tab:tab1}
\end{center}
\end{table*}

\begin{table*}[t]
\caption{Comparison with the state-of-the-art methods on REVERIE dataset.}
\begin{center}
  \resizebox{\linewidth}{!}{
  \begin{tabular}{l|rrrrrr|rrrrrr|rrrrrr}

  \hline
  \multicolumn{1}{c}{\multirow{3}{*}{Methods}}  & \multicolumn{6}{|c|}{REVERIE Validation Seen} &\multicolumn{6}{c|}{REVERIE Validation Unseen} & \multicolumn{6}{c}{REVERIE Test Unseen} \\
  \cline{2-19}& \multicolumn{4}{c|}{Navigation}  & \multicolumn{1}{c}{\multirow{2}{*}{RGS$\uparrow$}}&\multicolumn{1}{c|}{\multirow{2}{*}{RGSPL$\uparrow$}} & \multicolumn{4}{c|}{Navigation}  & \multicolumn{1}{c}{\multirow{2}{*}{RGS$\uparrow$}}&\multicolumn{1}{c|}{\multirow{2}{*}{RGSPL$\uparrow$}} & \multicolumn{4}{c|}{Navigation}   & \multicolumn{1}{c}{\multirow{2}{*}{RGS$\uparrow$}}&\multicolumn{1}{c}{\multirow{2}{*}{RGSPL$\uparrow$}} \\
  \cline{2-5} \cline{8-11} \cline{14-17}  & \multicolumn{1}{c}{SR$\uparrow$} & \multicolumn{1}{c}{OSR$\uparrow$} & \multicolumn{1}{c}{SPL$\uparrow$}  & \multicolumn{1}{c|}{TL} &  &  & \multicolumn{1}{c}{SR$\uparrow$} & \multicolumn{1}{c}{OSR$\uparrow$} & \multicolumn{1}{c}{SPL$\uparrow$} &\multicolumn{1}{c|}{TL} & & & \multicolumn{1}{c}{SR$\uparrow$} & \multicolumn{1}{c}{OSR$\uparrow$} & \multicolumn{1}{c}{SPL$\uparrow$} & \multicolumn{1}{c|}{TL} &  & \\
  \hline 
  Random &2.74 &8.92& 1.91 & \multicolumn{1}{c|}{11.99} & 1.97 & 1.31 &1.76& 11.93 &1.01& \multicolumn{1}{c|}{10.76}  & 0.96 & 0.56 &  2.30  &8.88 & 1.44&  \multicolumn{1}{c|}{10.34} &1.18 & 0.78 \\
  Human & -- & --  & -- & \multicolumn{1}{r|}{--} & -- & -- & -- & --  & --  & \multicolumn{1}{r|}{--} & -- & -- & 81.51 & 86.83  & 53.66 & \multicolumn{1}{r|}{21.18} & 77.84 &  51.44 \\
  \hline
  Seq2Seq-SF \cite{anderson2018vision} &29.59&35.70 & 24.01 & \multicolumn{1}{r|}{12.88}  &18.97& 14.96 & 4.20& 8.07 & 2.84   & \multicolumn{1}{r|}{11.07}  & 2.16 & 1.63 & 3.99 & 6.88  & 3.09 & \multicolumn{1}{r|}{10.89} & 2.00 & 1.58 \\
  RCM \cite{wang2019reinforced} & 23.33 & 29.44 & 21.82& \multicolumn{1}{r|}{10.70} & 16.23 & 15.36 & 9.29 & 14.23 & 6.97 &\multicolumn{1}{r|}{11.98} & 4.89 & 3.89 & 7.84 & 11.68 & 6.67 & \multicolumn{1}{r|}{10.60} & 3.67 & 3.14\\
  SMNA \cite{DBLP:conf/iclr/MaLWAKSX19} & 41.25& 43.29  & 39.61& \multicolumn{1}{r|}{7.54}   & 30.07 &  28.98 & 8.15 & 11.28 &6.44 & \multicolumn{1}{r|}{9.07} & 4.54& 3.61 & 5.80& 8.39 & 4.53 &\multicolumn{1}{r|}{9.23}   & 3.10& 2.39 \\
  FAST-Short \cite{ke2019tactical} & 45.12& 49.68 &40.18& \multicolumn{1}{r|}{13.22}  &31.41 & 28.11 & 10.08 & 20.48 & 6.17 & \multicolumn{1}{r|}{29.70}  & 6.24 & 3.97 & 14.18 & 23.36 & 8.74 & \multicolumn{1}{r|}{30.69}  & 7.07 & 4.52 \\
  FAST-MATTN \cite{qi2020reverie} & \high{50.53} & \high{55.17} & \high{45.50} & \multicolumn{1}{r|}{{16.35}} & \high{31.97} & \high{29.66} & \ 14.40 & \high{28.20} &  7.19 & \multicolumn{1}{r|}{{45.28}}  & \ 7.84 & \ 4.67 & \ 19.88 & \ 30.63 & \ 11.61 & \multicolumn{1}{r|}{{39.05}} & \ 11.28 & \ 6.08 \\
  ProbES~\cite{DBLP:conf/acl/LiangZLXL22} & \ 46.52 & \ 48.49 & \ 42.44 & \multicolumn{1}{r|}{{13.59}} & \ 33.66 & \ 30.86 & \high{27.63} & \ 33.23 & \high{22.75} & \multicolumn{1}{r|}{{18.00}} & \high{16.84} & \high{13.94} & \high{24.97} & \ 28.23 & \high{20.12} & \multicolumn{1}{r|}{{17.43}} & \high{15.11} & \high{12.32} \\
  \hline
  RecBERT (init. OSCAR) \cite{DBLP:conf/cvpr/Hong0QOG21}& \ 39.85 & \ 41.32 & \ 35.86 & \multicolumn{1}{r|}{{12.85}} & \ 24.46 & \ 22.28 & \high{25.53} & \ 27.66 & \high{21.06} & \multicolumn{1}{r|}{{14.35}} & \high{14.20} & \high{12.00} & \high{24.62} & \ 26.67 & \high{19.48} & \multicolumn{1}{r|}{{14.88}} & \high{12.65} & \high{10.00} \\
  RecBERT (init. PREVALENT) \cite{DBLP:conf/cvpr/Hong0QOG21} & \ 51.79 & \ 53.90 & \ 47.96 & \multicolumn{1}{r|}{{13.44}} & \ 38.23 & \ 35.61 & \high{30.07} & \ 35.02 & \high{24.90} & \multicolumn{1}{r|}{{16.78}} & \high{18.77} & \high{15.27} & \high{\textbf{29.61}} & \ 32.91 & \high{\textbf{23.99}} & \multicolumn{1}{r|}{{15.86}} & \high{16.05} & \high{\textbf{13.51}} \\
  \hline
  PANDA (Ours) & \textbf{54.39} & \textbf{55.80} & \textbf{51.08} & \multicolumn{1}{r|}{{13.04}} & \textbf{40.62} & \textbf{38.29} & \high{\textbf{32.66}} & \textbf{37.66} & \high{\textbf{27.88}} & \multicolumn{1}{r|}{15.74} & \high{\textbf{20.76}} & \high{\textbf{17.74}} & \high{27.81} & \textbf{33.30} & \high{20.89} & \multicolumn{1}{r|}{18.30} & \high{\textbf{17.20}} & \high{12.95} \\
  \hline 
\end{tabular}}
\end{center}
\vspace{-0.4cm}
\label{tab:table2}

\end{table*}

\subsection{Comparison to State-of-the-Art Methods}

\noindent
\textbf{Results on R2R.} 
Results in Table \ref{tab:tab1} compare the performance of different methods on the R2R. Compared to the baseline model RecBERT \cite{DBLP:conf/cvpr/Hong0QOG21}, PANDA improves the agent’s performance, achieving 66\% SR \textbf{(+3\%)} and 60\% SPL \textbf{(+3\%)} on the validation unseen. On the test unseen split, we achieve 64\% SR \textbf{(+1\%)} and 59\% SPL \textbf{(+2\%)}. The large performance improvement suggests that improving the indoor-aware and context-aware capacity for pretrained models benefits the learning of navigation for the VLN agent. Compared to existing state-of-the-art methods, we can see that only 2\% performance gap on SR exists between the validation unseen and the test unseen splits, indicating that our agent improves the generalization ability to new environments. Among all the methods, PANDA has the best results across all metrics, even compared against some newest entries such as SEvol \cite{DBLP:conf/cvpr/ChenGMZ022}, ProbES \cite{DBLP:conf/acl/LiangZLXL22} and ADAPT \cite{DBLP:conf/cvpr/Lin0CLLL22}. Notice that the counterpart methods, with data augmentation or better navigation inference, are orthogonal to the proposed PANDA, meaning that they can be integrated to yield even stronger solutions.

\noindent
\textbf{Results on REVERIE.}
We compare PANDA with existing state-of-the-art methods on the REVERIE dataset, as shown in Table \ref{tab:table2}. Compared to the baseline model RecBERT (init. PREVALNRT) \cite{DBLP:conf/cvpr/Hong0QOG21}, we achieve 1.99\% improvement on RGS and 2.47\% improvement on RGSPL on the validation unseen split. On the test unseen split, we achieve 1.15\% improvement on RGS. Despite RecBERT having a higher SR and SPL, one of the possible reasons is that their agent is wandering in the process of finding the target object. This suggests that PANDA is better for locating target objects.

\subsection{Ablation Study}
\label{Ablation Study}
In this subsection, we conduct experiments to further study the effectiveness of the prompting stages in PANDA, verify the effects of visual prompts based on

\noindent
different pretrained models, and subsequently investigate the effects of different design choices in the context-aware stage.

\vspace{-0.6cm}
\begin{table}
\small
\centering
\setlength\tabcolsep{4.5pt}
\caption{Ablation study of different prompt forms on R2R.}
\label{tab: tab3}	
\renewcommand\arraystretch{1.0}
\begin{tabular}{ccccccc} \toprule
\multirow{2}{*}{Methods}&\multicolumn{3}{c}{Val seen}&\multicolumn{3}{c}{Val Unseen}\cr
&NE  $\downarrow$ &SR $\uparrow$ &SPL $\uparrow$ &NE $\downarrow$ &SR $\uparrow$ &SPL $\uparrow$\cr
\hline

RecBERT (init. PREVALENT) \cite{DBLP:conf/cvpr/Hong0QOG21}&2.90&72.18&67.72&3.93&62.75&56.84\\
\hline
+ Deep visual prompts&\textbf{2.31}&\textbf{76.49}&71.70&3.72&\ 65.05 &\ 59.33 \\ 
+ Hard context prompts&2.65&74.14&70.04&\ 3.78 &64.84&59.41\\ 
\hline
PANDA&2.54&\ 75.02&\textbf{71.84}&\textbf{3.50}&\textbf{65.60}&\textbf{59.71}\\ 
\hline

\vspace{-0.6cm}
\end{tabular}
\end{table}

\noindent
\textbf{Overall Effectiveness of the Two Prompting Stages.} Table \ref{tab: tab3} shows the comparison of using different prompt forms on the R2R dataset. Introducing deep visual prompts and hard context prompts can effectively improve the performance of the strong baseline model RecBERT (init. PREVALENT) \cite{DBLP:conf/cvpr/Hong0QOG21}. By comparing the results between the baseline and only with deep visual prompts, we can find that deep visual prompts can effectively enhance navigation performance, demonstrating that deep visual prompts with additional knowledge for visual recognition are useful for understanding indoor image semantics. Comparing the results between the baseline and only with hard context prompts, we can see that the introduction of the hard context prompts improves the navigation performance, which shows that attending to the contextual relations between actions and the sequential cross-modal alignment is helpful for making corrective action decisions. By comparing the results between only with deep visual prompts and PANDA, we can find that introducing hard context prompts can further improve navigation performance.

\noindent
\textbf{Effects of Different Contrast Learning Objectives
.} 
Table \ref{tab: tab3} shows the effectiveness of hard prompts used in the context-aware stage where we incorporate three key components into the total loss function: count, individual, and overall losses. We have evaluated the efficacy of each loss component, and Table \ref{tab: tab5} presents the quantitative results, revealing the positive impact of all three losses on the final performance.

\begin{table}
\small
\centering
\setlength\tabcolsep{4.5pt}

\caption{Ablation study of the context-aware stage.}
\label{tab: tab5}
		
{\renewcommand{\arraystretch}{1.0}}

\begin{tabular}{ccccccc}

\specialrule{.05em}{.05em}{.05em}

\multirow{2}{*}{Methods}&\multicolumn{3}{c}{Val seen}&\multicolumn{3}{c}{Val Unseen}\cr

&NE  $\downarrow$ &SR $\uparrow$ &SPL $\uparrow$ &NE $\downarrow$ &SR $\uparrow$ &SPL $\uparrow$\cr
\hline

RecBERT (init. PREVALENT) \cite{DBLP:conf/cvpr/Hong0QOG21}&2.90&72.18&67.72&3.93&62.75&56.84\\
\hline
+ $\mathcal{L}_{sub}$ & 2.73 & \textbf{74.22} & 68.51 & 3.88 & 63.14 & 57.23 \\
\hline

+ $\mathcal{L}_{cnt}$ &2.86& 72.92 & 68.52 & 3.91 & 63.37 & 57.92 \\
+ $\mathcal{L}_{cnt}$+$\mathcal{L}_{ind}$ &2.71& 73.57 & 69.43 & 3.83 & 64.26 & 58.64 \\
+ $\mathcal{L}_{cnt}$+$\mathcal{L}_{ind}$+$\mathcal{L}_{ove}$&\textbf{2.65}& 74.14 & \textbf{70.04} & \textbf{3.78} & \textbf{64.84} & \textbf{59.41} \\
        
    \specialrule{.05em}{.05em}{.05em}
\end{tabular}
\end{table}

In order to identify whether it is the context prompting or merely further exposure to the target domain that actually provides an improvement, we remove all hard context prompts and just match sub-instructions with sub-paths by contrastive learning. This is simply represented as $\mathcal{L}_{sub}$. Through a comparison of the experimental results presented in Table \ref{tab: tab5}, we observe that aligning only sub-instructions and sub-paths yields only a marginal improvement in navigation performance by merely considering the out-of-context cross-modal alignment, while incorporating hard context prompts is proved to be more effective at enhancing performance by explicitly capturing sequence-level semantics and performing a sequential cross-modal alignment. Obviously, $\mathcal{L}_{sub}$ has even lower performance than only $\mathcal{L}_{cnt}$ on the validation unseen split even though $\mathcal{L}_{cnt}$ only considers the number of navigation steps, which is a very general contextual hint. With $\mathcal{L}_{cnt}$ and $\mathcal{L}_{ind}$ combined, the gap becomes much larger. VLN is a task characterized by sequential decision-making, where not only out-of-context knowledge is important but also contextual relations. This also suggests that hard context prompts are capable of capturing higher-order relationships among navigational actions at the sequence level.

\section{Conclusion}
In this work, we propose a Prompt-bAsed coNtext- and inDoor-Aware (PANDA) pretraining framework, which prompts the VLN agent with the capability of recognizing objects and scenes in visual perceptions in the indoor-aware stage. In the context-domain stage, PANDA enables a sequence-level representation via hard context prompts that are aware of the semantics of individual image-text pairs and across navigational actions along the trajectory. The context prompts in this work dig into the potential of prompt-based learning approaches for understanding ordinal actions and contextual relations. We believe that PANDA also can benefit future studies in other vision and language tasks. We will leave this for future work.

\section*{Acknowledgement}

This research was supported partially by the National Natural Science Fund of China (Grant Nos. 62306329 and 62103425, and the Natural Science Fund of Hunan Province (Grant Nos. 2023JJ40676 and 2022JJ40559).

%
%
%
%




\clearpage
\bibliographystyle{ieeetr} 
\bibliography{sample-base}

\end{document}